\documentclass[conference]{IEEEtran}
\IEEEoverridecommandlockouts
\usepackage{cite}
\usepackage[numbers]{natbib}
\usepackage{amsmath,amssymb,amsfonts}
\usepackage{algorithm}
\usepackage{algorithmic}
\usepackage{graphicx}
\usepackage{textcomp}
\usepackage{xcolor}
\usepackage{booktabs}
\usepackage{multirow}
\usepackage{subfigure}
\usepackage{soul}
\usepackage{url}
\usepackage{pifont} 
\usepackage[colorlinks,
            linkcolor=red,
            anchorcolor=blue,
            citecolor=green
            ]{hyperref}
\usepackage{caption}
\def\BibTeX{{\rm B\kern-.05em{\sc i\kern-.025em b}\kern-.08em
    T\kern-.1667em\lower.7ex\hbox{E}\kern-.125emX}}
\renewcommand\thesection{\arabic{section}}
\renewcommand\thesubsectiondis{\thesection.\arabic{subsection}}
\renewcommand\thetable{\arabic{table}}

\begin{document}

\newcommand{\notice}[1]{\textcolor{blue}{#1}}

\def\arraystretch{0.60}

\title{On the Multi-modal Vulnerability of Diffusion Models}

\author{\IEEEauthorblockN{Dingcheng Yang$^{1,*}$, Yang Bai$^{2,*}$, Xiaojun Jia$^3$, Yang Liu$^3$, Xiaochun Cao$^4$, Wenjian Yu$^{1,\dag}$\thanks{$^*$Equal contribution. $^{\dag}$Corresponding author.}}
\IEEEauthorblockA{$^1$Dept. Computer Science \& Tech., BNRist, Tsinghua University, Beijing, China.  \\
$^2$Independent Researcher.\\
$^3$Nanyang Technological University.\\
$^4$Sun Yat-sen University, Shenzhen.\\
ydc19@mails.tsinghua.edu.cn, baiyang0522@gmail.com, jiaxiaojunqaq@gmail.com, \\
yangliu@ntu.edu.sg, caoxiaochun@mail.sysu.edu.cn, yu-wj@tsinghua.edu.cn
}
}

\maketitle

\begin{abstract}

Diffusion models have been widely deployed in various image generation tasks, demonstrating an extraordinary connection between image and text modalities. Although prior studies have explored the vulnerability of diffusion models from the perspectives of text and image modalities separately, the current research landscape has not yet thoroughly investigated the vulnerabilities that arise from the integration of multiple modalities, specifically through the joint analysis of textual and visual features. In this paper, we are the first to visualize both text and image feature space embedded by diffusion models and observe a significant difference. The prompts are embedded chaotically in the text feature space, while in the image feature space they are clustered according to their subjects. These fascinating findings may underscore a potential misalignment in robustness between the two modalities that exists within diffusion models. Based on this observation, we propose MMP-Attack, which leverages multi-modal priors (MMP) to manipulate the generation results of diffusion models by appending a specific suffix to the original prompt. Specifically, our goal is to induce diffusion models to generate a specific object while simultaneously eliminating the original object. Our MMP-Attack shows a notable advantage over existing studies with superior manipulation capability and efficiency.  Our code is publicly available at \url{https://github.com/ydc123/MMP-Attack}.
\end{abstract}

\section{Introduction}

In recent years, diffusion models~\cite{ho2020denoising,song2020denoising} have revolutionized the field of image generation, achieving state-of-the-art results in both the diversity and quality of generated content. The advancement of vision-language models~\cite{radford2021learning} has further enhanced the capabilities of diffusion models, giving rise to novel applications in text-to-image (T2I) generation~\cite{rombach2022high,ramesh2022hierarchical,nichol2022glide,saharia2022photorealistic,gal2023an}. However, existing studies have shown that diffusion models also exhibit vulnerability issues, where appending a specific suffix to the original prompts can manipulate diffusion models to generate completely different image content. ATM~\cite{du2024stable} and SAGE~\cite{liu2023intriguing} have respectively investigated white-box attacks for both untargeted and targeted scenarios. Additionally, Adversarial Prompting (AP)~\cite{maus2023black} explored query-based targeted attack. However, all three methods demand a significant number of image generations, not only making it time-consuming but also unsuitable for commercial models due to their confidentiality or substantial monetary costs. QF-Attack~\cite{zhuang2023pilot} introduces an attack method that does not require image generation, but it is limited to untargeted attack and targeted erasing, namely generating random image content unrelated to the original prompt and omitting a specific category mentioned in the original prompt respectively. MMA-Diffusion~\cite{yang2024mma} optimized adversarial prompts to make T2I models generate Not-Safe-For-Work (NSFW) images. However, it did not consider the issue from a robustness perspective, meaning it did not require modifications on a given original prompt. Consequently, the problem addressed by MMA-Diffusion was relatively easier. 


\begin{figure}[t]
  \centering
  \subfigure[Visualization of text features.]
    {
    \includegraphics[width=1.5in]{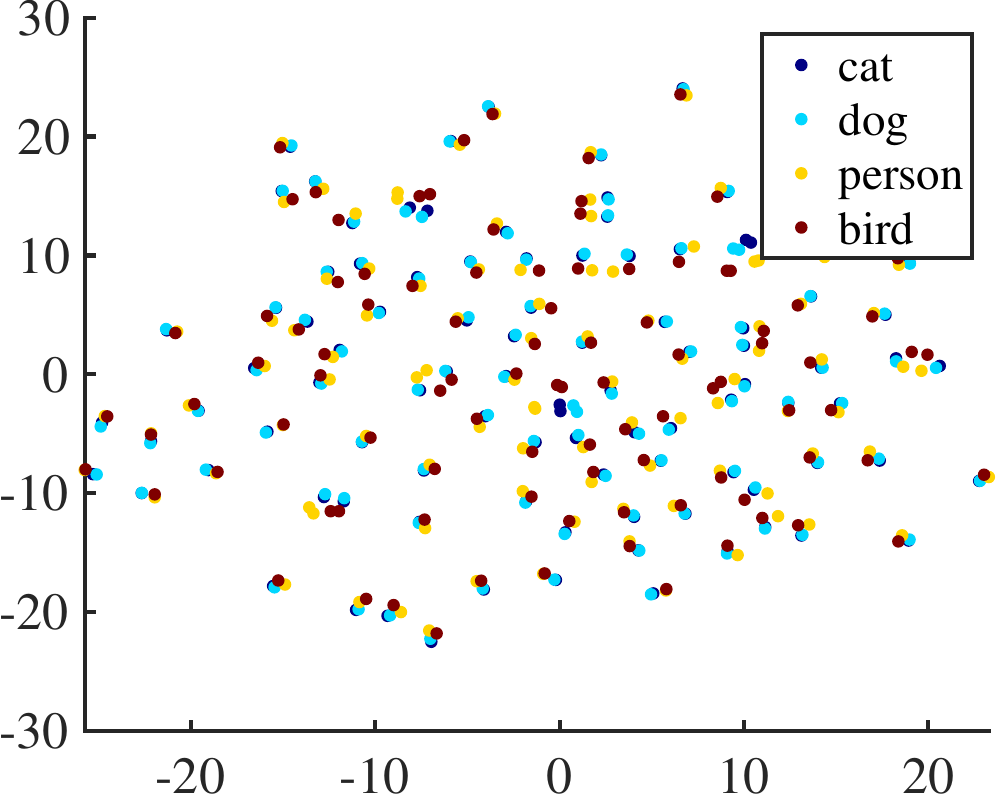}
    \label{fig:text}
    }
    \subfigure[Visualization of image features.]
    {
    \includegraphics[width=1.5in]{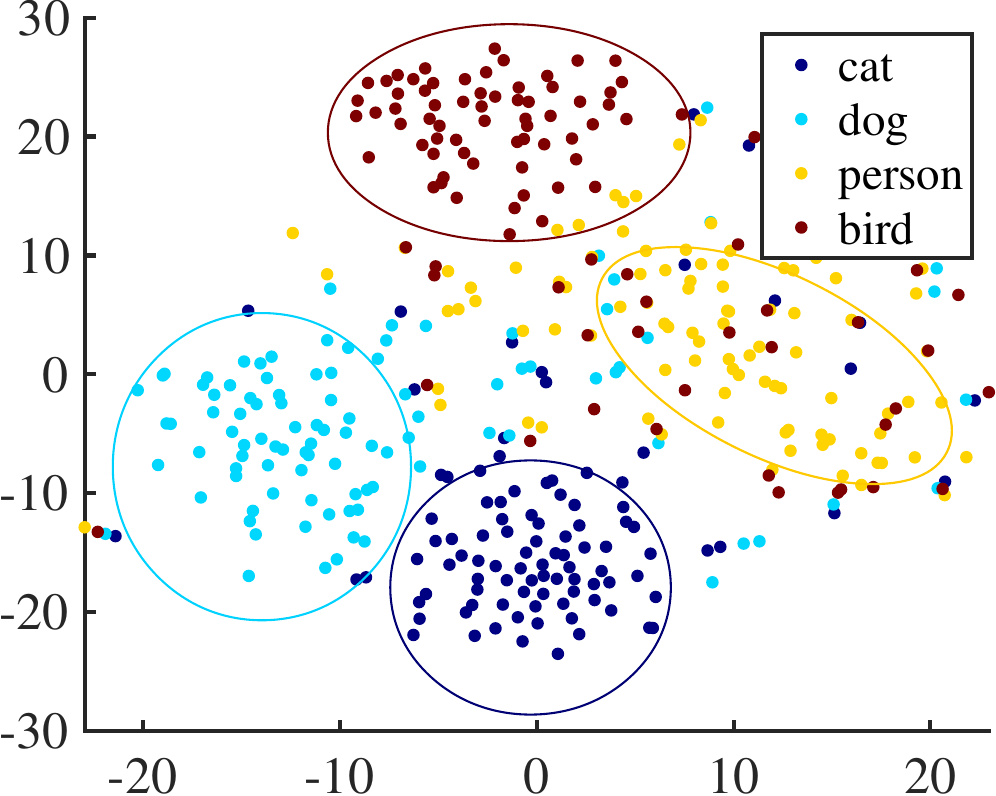}
    \label{fig:img}
    }
    \caption{Visualization of 400 samples in text (left) and image (right) feature space embedded by Stable Diffusion v1.4 (SD v14). Text features are chaotic while image features are clustered.}
\label{fig:intro}
\end{figure}

Besides their unsatisfactory performance, existing studies are also limited by their designed algorithms solely on either text or image feature space. In QF-Attack and MMA-Diffusion, the distance from the original prompt and target prompt was calculated in the text feature space. In ATM, SAGE and AP, the objective functions were designed in the image feature space with an auxiliary image classifier, assessing whether the generated images contain objects of predefined categories. The lack of exploration across different modalities inspired us to visualize the text and image feature space within diffusion models simultaneously. A significant difference between the two modalities is thus observed. As shown in Fig.~\ref{fig:intro}, we draw both text and image features of 400 samples embedded by Stable Diffusion v1.4 (SD v14), which are formed with 100 templates and 4 objects. Details of these samples are given in Sec. 3.2. The prompts are embedded chaotically in the text feature space, while in the image feature space they are clustered according to their subjects. This phenomenon can be attributed to the fact that text features distribute their emphasis across a variety of words, consequently placing greater importance on the sentences or templates. In contrast, image features are more concentrated on the specific objects. This difference also highlights certain suboptimal alignments within existing diffusion models, particularly from a robustness standpoint, which we will further analyse in Sec. 3.2.

Based on this observation, we propose an \textbf{MMP-Attack} by utilizing \textbf{M}ulti-\textbf{M}odal \textbf{P}riors. Our approach optimizes a suffix appended to the original prompt, aiming to effectively manipulate T2I diffusion models. Specifically, it facilitates the generation of a desired target object by removing the original object, thus addressing the most challenging scenario. The distance between the optimized prompt and the target category (to add) is minimized in both text and image feature space. The differences between our MMP-Attack and existing works are briefly summarized in Table~\ref{tab:setting}.

\begin{table}[h]
    \centering
    \caption{Comparison of existing methods with ours, based on the considered modality, targeted/untargeted setting, and whether image generation is required.}
    \label{tab:setting}

    \resizebox{0.45 \textwidth}{!}
    { 
    \begin{tabular}{@{}c|c@{~~}|c@{~~}|c}
        \toprule
         Method & Modality & Goal  & Generation-free\\
         \midrule
         ATM & image & Untargeted attack & \ding{53}\\
         SAGE&  image & Targeted attack & \ding{53}\\
         AP& image & Targeted attack & \ding{53}\\
         QF-Attack&   text & Untargeted attack & \ding{51}\\
         MMA-Diffusion & text & NSFW & \ding{51}\\
         \midrule
         \textbf{MMP (Ours)} & text+image & Targeted attack & \ding{51}\\
         
         \bottomrule
    \end{tabular}
    }

\end{table}

Comprehensive experiments demonstrate the superior universality and transferability of our optimized suffix. Universality indicates that a suffix optimized under a specific prefix can, to some extent, generalize to other prefixes. Transferability indicates that the suffix optimized on an open-source diffusion model can be employed to manipulate a black-box diffusion model, posing a more severe security threat to commercial T2I models, such as DALL-E 3. In this paper, our experimental results show an attack success rate of 50.4\% in manipulating Stable Diffusion v2.1 (SD v21) using the suffix generated on SD v14. Moreover, after analyzing the optimized suffix, we observed that MMP-Attack often works in a \textbf{\textit{cheating}} way, which means that it often contains some tokens related to the target object. \textit{It should be noted that simply appending the target object to the original prompt does not work.} Therefore, we also denote the suffix we optimize as a \textbf{\textit{cheating suffix}}.

The major contributions are summarized as follows.
\begin{itemize}
    \item We conduct a visual analysis of both the text and image feature spaces that are embedded by diffusion models. Our work represents the first instance of observing the notable differences in features across multi-modalities. Such intriguing observations could potentially highlight a misalignment between the two modalities within diffusion models, particularly from the perspective of robustness.
    \item Based on the observations, we propose \textbf{MMP-Attack}, which leverages multi-modal priors (MMP) to targeted manipulate the generation results of diffusion models. This is achieved by appending a specific suffix after the original prompt, which often contains some tokens related to the target object, hence referred to as a \textit{cheating suffix}. 

    \item Experimental results indicate that our method achieves over 81.8\% attack success rates on two open-source T2I models even with only four tokens, showcasing a notable advantage over existing works. 
\end{itemize}

\section{Related Work}
\subsection{Diffusion Models}
Diffusion models have achieved remarkable success in the field of image generation through a learnable step-wise denoising process that transforms a simple Gaussian distribution into the data distribution~\cite{ho2020denoising}. Some studies have been proposed to accelerating the image generation process~\cite{song2020denoising,lu2022dpm}. Beyond the field of image synthesis, diffusion models are widely used in diverse fields, including music generation~\cite{huang2023noise2music}, 3D generation~\cite{wang2023prolificdreamer} and video generation~\cite{blattmann2023stable}. Notably, by combining with the visual language model CLIP~\cite{radford2021learning}, the diffusion model showcases exceptional prowess in text-to-image generation~\cite{rombach2022high,gal2023an}.

\subsection{Manipulation in T2I Generation}
Deep neural networks are known to be vulnerable~\cite{szegedy2013intriguing,zhao2021success,yang2023boosting,yang2023generating,bai2020improving,bai2023query}. Recent studies have shown that the T2I generation process is vulnerable to prompts, indicating that it is possible to manipulate T2I models to generate images unrelated to the given prompt by adding a special suffix to the prompt. ATM~\cite{du2024stable} and SAGE~\cite{liu2023intriguing} proposed white-box methods, which assume that the diffusion model is fully known, making it unsuitable for confidential commercial models. AP~\cite{maus2023black} performed a high-cost query-based method. The practicality of these approaches is limited. QF-Attack~\cite{zhuang2023pilot} assumed that the diffusion model has a white-box CLIP model but an inaccessible and unqueryable generative model. Under this assumption, they proposed a generation-free method against T2I models, which employed a genetic algorithm to manipulate the CLIP model. However, they only considered untargeted attack and targeted erasing. In this paper, we follow the setting outlined in QF-Attack but address a more challenging task: targeted manipulation, specifically by adding target objects while removing original objects in original prompts. 

Another line of work in manipulating T2I models involves designing prompts to make the generated content NSFW, such as with MMA-Diffusion~\cite{yang2024mma}. The key difference between MMP-Attack and such approaches is the presence of the original prompt, which makes the problem we are studying significantly more difficult. The algorithm of MMA-Diffusion can be modified to generate \textit{cheating suffixes}, and a comparative experiment with MMP-Attack will also be presented in Sec. 5.2.

\section{Observations on Multi-modal Features within Diffusion Models}
In this part, we first outline the preliminary of diffusion models. We then define and visualize the text and image feature spaces, thereby highlighting a marked distinction between the multi-modal feature distributions.

\subsection{Preliminary: Pipeline of Diffusion Model}

Given that the vocabulary of candidate tokens forms a set $\mathbb{V}=\{w_1,w_2,\cdots,w_{|\mathbb{V}|}\}$, an input prompt can be expressed as $s \in \mathbb{V}^*$. A well-trained diffusion model consists of two components: a CLIP model and a generative model $G$. The CLIP model includes an image encoder $F^i$, which takes an image as input and outputs a $d_{\text{emb}}$-dimensional image embedding vector. It also includes a token embedder $E_{\psi}$ and a text encoder $F^t$, which together embed a text prompt into a $d_{\text{emb}}$-dimensional text embedding vector. Here, $\psi \in \mathbb{R}^{|\mathbb{V}| \times d_{\text{token}}}$ serves as an embedding codebook. For the input prompt $s$, $E_{\psi}(s)$ is a matrix of shape $|s| \times d_{\text{token}}$, where $E_{\psi}(s)_i = \psi_j$, with the condition that $w_j = s_i$. This token embedding matrix $E_{\psi}(s)$ is then input into the text encoder $F^t$ and embedded as a $d_{\text{emb}}$-dimensional text embedding vector. During the training stage, an image is transformed to an image embedding vector by the image encoder. Simultaneously, its caption (text data) is transformed to a text embedding vector by the token embedder and the text encoder. The distance between the two vectors is minimized to enable the CLIP model to align the image space and text space. During the T2I generation stage, the input prompt $s$ is first embedded into a text embedding vector $v$ by the token embedder and text encoder. Then, it is input into the subsequent generative model $G$ to sample $x \sim G(v)$, where $G(v)$ is a probability distribution conditioned on $v$, and $x$ represents a sampled image. Thus, the T2I generation from the input prompt $s$ can be understood as a process of sampling from the probability distribution $x \sim G(F^t(E_{\psi}(s)))$.

\subsection{Multi-modal Features in Diffusion Models}

Previous studies have separately investigated the vulnerability of diffusion models from the perspectives of text and image modalities~\cite{zhuang2023pilot,liu2023intriguing}. In contrast to their studies, we investigate the vulnerability of multi-modal features. Given a prompt $s$, we define its text embedding vector as $F^t(E_{\psi}(s))$, and its image embedding vector as $F^i(x)$, where $x \sim G(F^t(E_{\psi}(s)))$. 
Then, we visualize the text and image feature spaces, showcasing a marked distinction between the multi-modalities.

\begin{figure}[t]
  \centering
  \subfigure
    {
    \includegraphics[width=1.5in]{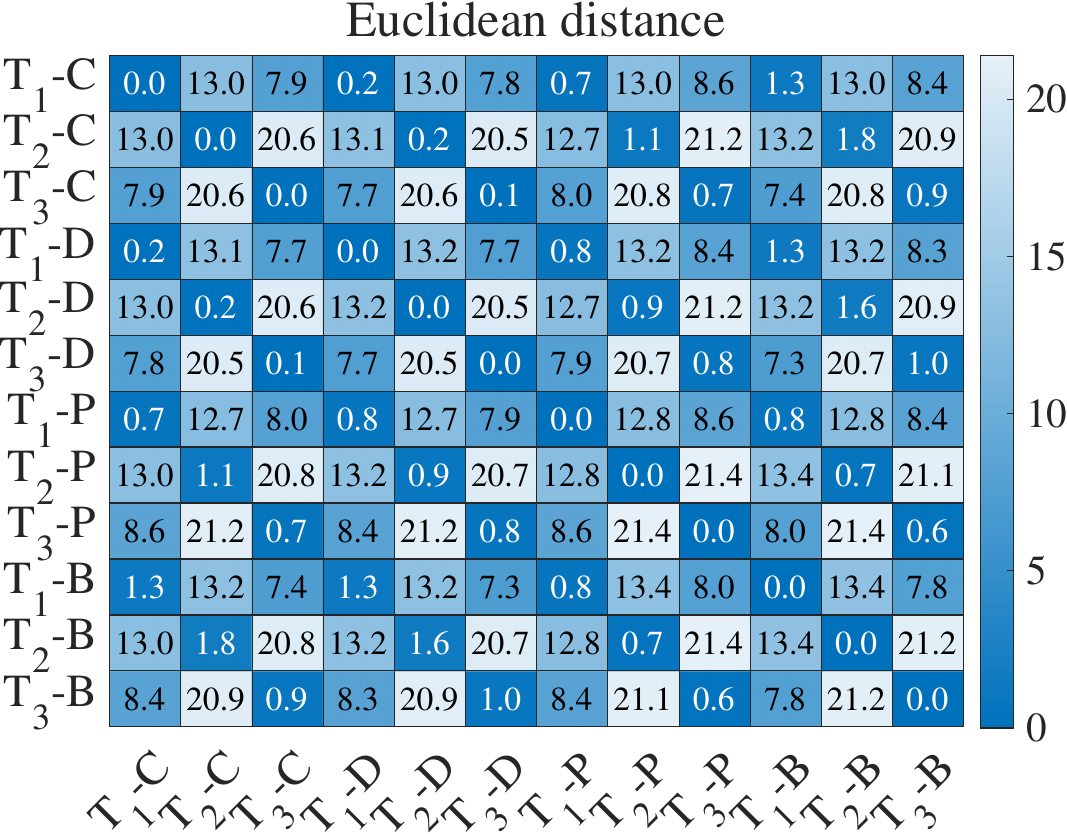}
    \label{fig:text-l2}
    }
    \subfigure
    {
    \includegraphics[width=1.5in]{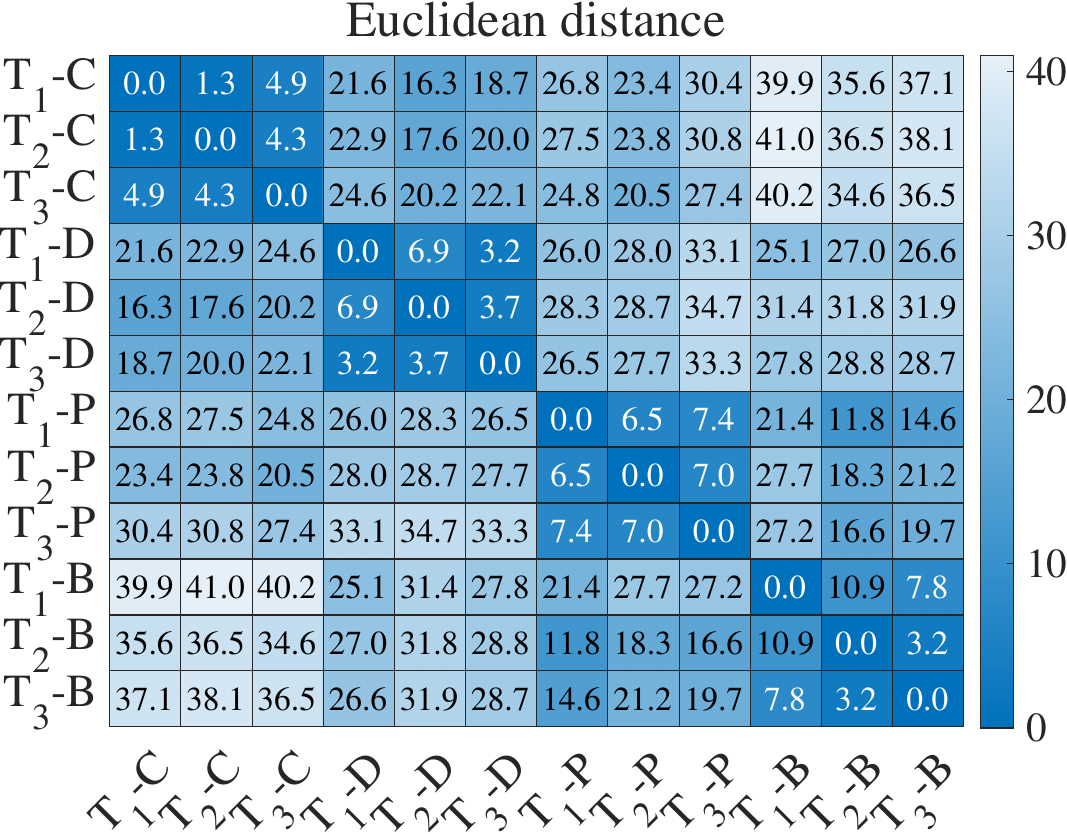}
    \label{fig:img-l2}
    }

    \caption{Euclidean distances between 12 different prompts in the text (left) and image (right) feature spaces. The prompts are generated from 3 different templates: `a \{noun\} is sitting on a bench in a park', `a \{noun\} is peeking out from behind a curtain', and `a \{noun\} is standing at the edge of a cliff', denoted as $T_1$, $T_2$, and $T_3$, respectively. `-C', `-D', `-P', and `-B' represent the \{noun\} being cat, dog, person, and bird respectively.}
\end{figure}

\textbf{Chaos Effect of Features in Text Space.}
We first visualize the text feature space. We instructed ChatGPT to generate 100 prompt templates, and then sequentially filled in `\texttt{cat}', `\texttt{dog}', `\texttt{bird}', and `\texttt{person}' sequentially as subjects to form 400 prompts. Then, we embedded these 400 prompts into text embedding vectors by the SD v14 and visualized them in the text feature space using t-SNE. The visualization results is shown in Fig.~\ref{fig:text}, illustrating that prompts associated with different subjects are mixed together. This is because both the subjects and other tokens are considered important by the text encoder. Thus, in the text feature space, prompts with different subjects but originating from the same template can be embedded close together. This implies that even if two prompts are close in the text feature space, they may have different subjects. To illustrate this phenomenon more clearly, we chose 12 prompts originating from 3 templates and calculated their Euclidean distances from each other, as shown in Fig.~\ref{fig:text-l2}.


\textbf{Clustering Effect of Features in Image Space.}
Then, we use SD v14 to visualize these 400 prompts on the image feature space, as shown in Fig.~\ref{fig:img}. It can be observed that the embedding vectors of these prompts have remarkably different distributions on the image feature space than on the text feature space, revealing the potential misalignment between text feature space and image feature space for diffusion models. Specifically, the prompts with the same subject are clustered together in the image feature space, while prompts with different subjects are distinguished from each other. This difference arises because the text encoder extracts features of all tokens in the prompts, while an image encoder primarily extracts features of the key object (subject) in the images. Thus, the prompts that are close in the image feature space often share the same subject. This can be evidenced in Fig.~\ref{fig:img-l2}, where the distances of prompt pairs with the same subject are significantly lower (up to 10.9) in the image feature space.

\section{Methodology}

\begin{figure}[h]
  \centering
    \includegraphics[width=3.3in]{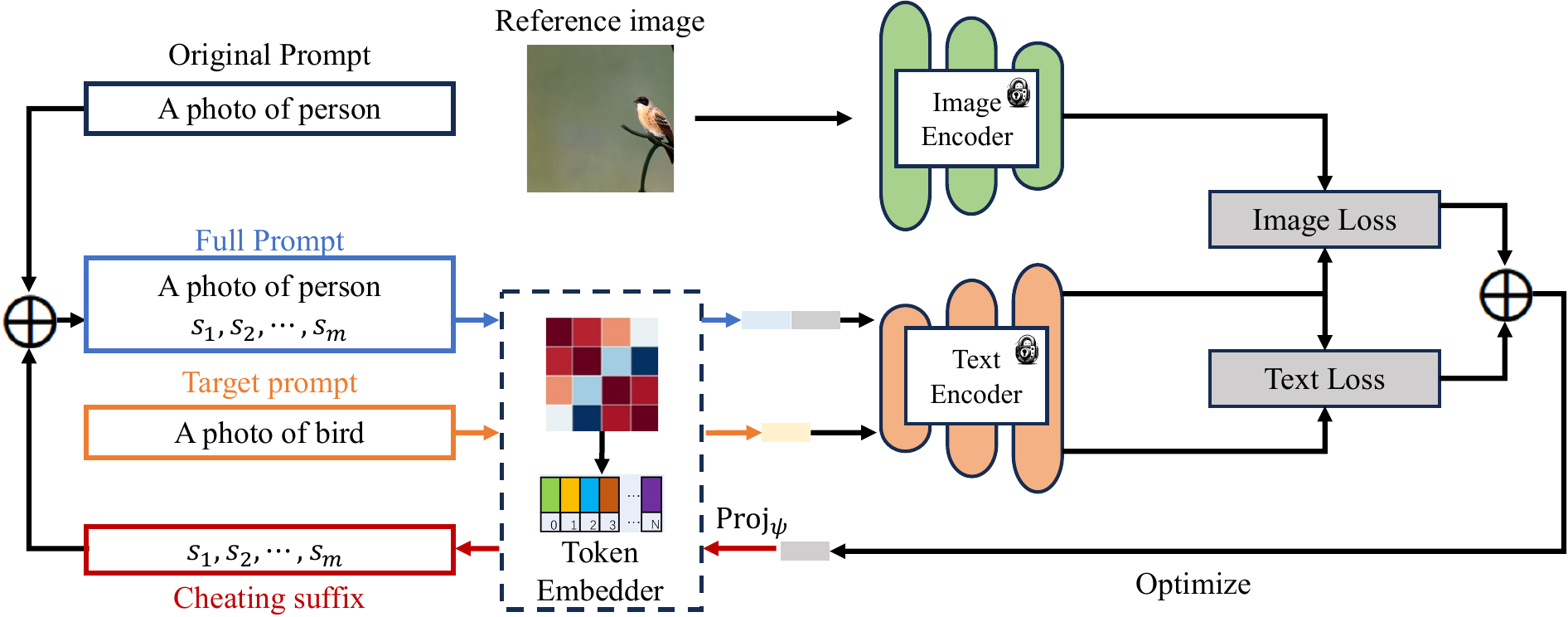}
    \caption{An illustration of the proposed MMP-Attack flow.}
      \label{fig:pipeline}
\end{figure} 

In this part, we propose MMP-Attack, which leverages multi-modal priors to targeted attack the T2I generation. We begin by formulating the targeted attack problem for T2I models. Then, motivated by the misalignment phenomenon observed in Sec. 3, we propose an optimization objective that simultaneously considers both the image and text modalities. Finally, we present the corresponding optimization approach. An illustration of our MMP-Attack is shown in Fig.~\ref{fig:pipeline}.

\subsection{Problem Formulation}
Let $s_o \in \mathbb{V}^n$ be the original prompt containing $n$ tokens, and $m$ be the number of tokens in the cheating suffix. The cheating suffix to be optimized can be represented as $s_a \in \mathbb{V}^m$, which will be concatenated with $s_o$ to get the full prompt $s_o \oplus s_a \in \mathbb{V}^{n+m}$, where the operator $\oplus$ denotes concatenation operator. 

For conducting targeted attack, we assume that there is a target category $t \in \mathbb{V}$ (e.g., \texttt{dog}, \texttt{bird}). This target category is irrelevant to the original prompt $s_o$. We need to search for a cheating suffix that, when concatenated with the original prompt $s_o$, guides the T2I diffusion model to generate an image containing the target category but is unrelated to $s_o$. The optimization objective is as follows:
\begin{equation}
\begin{aligned}
    \text{argmax}_{s_a} \mathbb{E}_{x\sim G(F^t(E_{\psi}(s_o \oplus s_a)))} \mathcal{A}(x,t,s_o)~,
\end{aligned}
    \label{eq:obj_ori}
\end{equation}
where $x$ represents a randomly generated image based on the full prompt $s_o \oplus s_a$. The $\mathcal{A}(x,t,s_o)$ is an evaluation metric to assess the manipulation performance. Following the assumptions of relevant work~\cite{zhuang2023pilot}, during the optimization process, we have access only to the CLIP model and are blind to the generative model $G$, which can only be used for evaluating the manipulation performance of the cheating suffix.

To ensure the naturalness of the cheating suffix, we refined the vocabulary $\mathbb{V}$ to include only English words that end with the `\texttt{</w>}' symbol, indicating a white-space. This step was necessary because the CLIP vocabulary includes tokens representing prefixes that do not end with `\texttt{</w>}', and the concatenation of such tokens could result in the optimized cheating suffix containing non-existent words, thereby reducing the naturalness of the prompt.
Furthermore, we additionally filtered out the top-20 synonyms of the target category from the vocabulary, to simulate real-world systems that block sensitive words.
 Specifically, the embedding codebook $\psi$ was employed to define the similarity between two tokens $w_i,w_j$ as $\cos(\psi_i,\psi_j)$, where $\cos(a,b)=\frac{a^Tb}{\|a\| \|b\|}$ represents the cosine similarity between two vectors. 

\subsection{Optimization Objective}
Directly solving ~\eqref{eq:obj_ori} is infeasible, because it involves a generative model $G$ that is unknown in our assumption. An alternative approach is to first construct a target vector $v_t$ that provides a favorable solution to the following optimization objective:
\begin{equation}
\begin{aligned}
    \text{argmax}_{v_t} \mathbb{E}_{x\sim G(v_t)} \mathcal{A}(x,t,s_o)~.
\end{aligned}
    \label{eq:obj_surrogate}
\end{equation}
Assuming such a $v_t$ exists, we can achieve a favorable solution to~\eqref{eq:obj_ori} by maximizing the similarity between the text embedding vectors of $s_o \oplus s_a$ and target vector $v_t$. Consequently, the optimization objective~\eqref{eq:obj_ori} is transformed into a simplified problem involving only $F^t$ and $E_{\psi}$:
\begin{equation}
\begin{aligned}
    \text{argmax}_{s_a} \cos(F^t(E_{\psi}(s_o \oplus s_a)),v_t).
\end{aligned}
    \label{eq:obj_clip}
\end{equation}

Although $G$ is unknown, constructing a favorable solution for problem~\eqref{eq:obj_surrogate} is not difficult, since we can use some heuristics solutions. For example, the images generated by a manually crafted prompt $s'=$`\texttt{a photo of} $t$' will undoubtedly satisfy the requirements of our targeted attack. Thus, we can utilize its text embedding vector $v_t^{text}=F^t(E_{\psi}(s'))$ as a target vector to guide the optimization of cheating suffix.

However, as demonstrated in Sec. 3, the embedding vectors of prompts in the text feature space and the image feature space are misaligned. This implies that even though prompts have relatively close distances in the text feature space, the resulting images could be far apart in the image feature space,  indicating differences in key objects present in the images. Thus, we integrate image modal information with text modal to guide the optimization process. Since the generative model $G$ is unknown, we cannot compute loss terms for generated images, as done in prior work~\cite{liu2023intriguing,maus2023black}. Instead, we propose a target vector based on image modality. This approach also avoids the costly image generation required in prior work that utilized the image modality. Specifically, given a reference image $x_t$ containing the target category, we calculate its image embedding vector $v_t^{image}=F^i(x_t)$, where $F^i$ is the image encoder of the CLIP model. The CLIP model possesses the characteristic that image-text pairs with higher correlation exhibit larger cosine similarities in their embedding vectors. Therefore, $v_t^{image}$ is also a favorable solution for~\eqref{eq:obj_surrogate}.  Finally, we concurrently optimize in both the image and text modalities, where the multi-modal loss is designed as maximizing the similarity between the text embedding vector of $s_o \oplus s_a$ and the two target vectors, as shown in Fig.~\ref{fig:pipeline}. The optimization objective is as follows:
\begin{equation}
\begin{aligned}
    \text{argmax}_{s_a} & \cos(v,v_t^{image})+\lambda \cos(v,v_t^{text}),\\
    & \text{s.t.} \quad v=F^t(E_{\psi}(s_o \oplus s_a)),
\end{aligned}
    \label{eq:loss}
\end{equation}
where $\lambda$ is a weighting factor to balance the loss terms between the image and text modalities.
\subsection{Optimization Approach}

The remaining challenge lies in solving~\eqref{eq:loss}. As the optimization variables are defined in a discrete space, it presents a combinatorial optimization problem that is  non-differentiable and often NP-hard. To address this issue, a commonly used technique is Straight-Through Estimation (STE) technique~\cite{bengio2013estimating}, which introduces a differentiable function sg$(\cdot)$ that is defined as the identity function during forward propagation and has zero partial derivatives. It has been previously applied in other discrete optimization domains, including neural network quantization~\cite{yang2021dynamic,yang2022dp}, and training discrete generative models such as VQ-VAE~\cite{van2017neural} and VQ-GAN~\cite{esser2021taming}. Inspired by these works, we leverage the sg$(\cdot)$ function to solve~\eqref{eq:loss}. Specifically, we optimize the token embedding matrix $Z \in \mathbb{R}^{m \times d_{\text{token}}}$ of the cheating suffix, and define a differentiable function $\text{Proj}_{\psi}: \mathbb{R}^{m \times d_{\text{token}}} \rightarrow \mathbb{R}^{m \times d_{\text{token}}}$, where $\text{Proj}_{\psi}(Z)_i=Z_i+ \text{sg}(\psi_j-Z_i)$ such that $j=\text{argmin}_{j'} \| \psi_{j'}-Z_i \|_2^2$. Notice that each row in matrix $\text{Proj}_{\psi}(Z)$ corresponds to an entry in the codebook $\psi$, therefore we can decode the cheating suffix $s_a=E_{\psi}^{-1}(\text{Proj}_{\psi}(Z))$. Moreover, due to the property $E_{\psi}(s_o \oplus s_a)=E_{\psi}(s_o) \oplus E_{\psi}(s_a)$,~\eqref{eq:loss} can be reformulated into the following optimization problem:
\begin{equation}
\begin{aligned}
    \text{argmax}_{Z} & \cos(v,v_t^{image})+\lambda \cos(v,v_t^{text}) \\
     \text{s.t.} \quad v&=F^t(E_{\psi}(s_o \oplus s_a))\\
     &=F^t(E_{\psi}(s_o \oplus E_{\psi}^{-1}(\text{Proj}_{\psi}(Z)))) \\
     &=F^t(E_{\psi}(s_o) \oplus \text{Proj}_{\psi}(Z)).
\end{aligned}
    \label{eq:obj_ste}
\end{equation}
Because the \text{Proj} function is differentiable,~\eqref{eq:obj_ste} can be solved using a gradient-based optimizer, providing better performance compared to prior work~\cite{zhuang2023pilot} that employs an zero-order optimizer. Our optimization approach is summarized in Algorithm~\ref{alg:opt}. The target conditional vectors are first calculated in Step 1-3. Then, the optimization variable $Z$ is initialized and optimized by a gradient descent algorithm (Step 4-13). Finally, the cheating suffix is decoded based on the optimized $Z$ (Step 14).

\begin{algorithm}[t]
\caption{MMP-Attack}\label{alg:opt}
\begin{flushleft}
\textbf{Input:} token embedder $E_{\psi}$, dimension of the token embedding vector $d_{\text{token}}$, text encoder $F^t$, image encoder $F^i$, learning rate $\eta$, number of iterations $N$, original prompt $s_o$, number of tokens in cheating suffix $m$, target category $t \in \mathbb{V}$, weighting factor $\lambda$, a reference image $x_t$ containing the target category $t$ and unrelated to original prompt $s_o$.\\
\textbf{Output:} Cheating suffix $s_a$.\\
\end{flushleft}
\begin{algorithmic}[1]
\STATE $v_t^{image} \gets F^i(x_t)$.
\STATE $s' \gets$ `\texttt{a photo of }$t$'.
\STATE $v_t^{text}=F^t(E_{\psi}(s'))$
\STATE Initialize $Z \in \mathbb{R}^{m \times d_{\text{token}}}$.
\STATE $bestloss \gets \infty, bestZ \gets Z$
\FOR{$i \gets 1$ to $N$}
\STATE $v \gets F^t(E_{\psi}(s_o) \oplus \text{Proj}_{\psi}(Z))$.
\STATE $\mathcal{L}=-\cos(v,v_t^{image})-\lambda \cos(v,v_t^{text})$. 
\IF{$bestloss > \mathcal{L}$}
    \STATE $bestloss \gets \mathcal{L}, bestZ \gets Z$.
\ENDIF
\STATE $Z \gets Z - \eta \nabla_Z \mathcal{L}$.
\ENDFOR

\STATE $s_a \gets E_{\psi}^{-1}(\text{Proj}_{\psi}(bestZ))$.
\end{algorithmic}
\end{algorithm}


A good initialization (Step 4) often helps reduce the complexity of the optimization problem, leading to better solutions~\cite{tashiro2020diversity}. To solve~\eqref{eq:obj_ste}, we consider three initialization methods:
\begin{enumerate}
    \item \textbf{EOS}: Initialize all $Z_i$ as the token embedding for \texttt{[eos]}, where \texttt{[eos]} is a special token in CLIP vocabulary representing the end of string. 
    \item \textbf{Random}: Randomly sample $m$ tokens from the filtered vocabulary and use their embeddings as the initial values for $Z$. 
    \item \textbf{Synonym}: Select the token with the highest cosine similarity to the target category $t$ in the filtered vocabulary, and use its token embedding as the initial values for all $Z_i$.
\end{enumerate}

The synonym initialization is used by default, and an ablation study on the initialization method is presented in Sec. 5.4.

\section{Experiments}
\subsection{Settings}
\textbf{Dataset.} Five object categories are selected from the Microsoft COCO dataset~\cite{lin2014microsoft}, namely \texttt{car}, \texttt{dog}, \texttt{person}, \texttt{bird}, and \texttt{knife}. They are considered as both the original and target categories, forming a total of $5 \times 4 = 20$ distinct category pairs. For each category pair, a cheating suffix is generated. Each cheating suffix is then used to generate $100$ images to evaluate the manipulation performance metrics. The final performance metrics are obtained by averaging across all categories, which means that for a given method, its performance metrics are calculated over $5 \times 4 \times 100 = 2000$ images.

\textbf{Models.} Following the setting in relevant work~\cite{zhuang2023pilot}, we initially employ Stable Diffusion v1.4 (SD v14)\footnote{https://huggingface.co/CompVis/stable-diffusion-v1-4} as the diffusion model for image generation and performance evaluation. This model utilizes a pretrained CLIP model, which is trained on a dataset containing text-image pairs~\cite{thomee2016yfcc100m}. Furthermore, we also consider an additional model, Stable Diffusion v2.1 (SD v21)\footnote{https://huggingface.co/stabilityai/stable-diffusion-2-1}, which has a distinct CLIP model compared to SD v14. During image generation, the resolution is set to $512 \times 512$, the number of inference step is set to 50, and the classifier-free guidance scale is set to 7.5. Finally, we also consider commercial T2I services, i.e., DALL-E 3~\cite{betker2023improving} and Imagine Art.


\textbf{Evaluation Metrics.} Given a generated image, the following metrics are considered to evaluate the manipulation performance: 1) \textbf{CLIP score}: We use the CLIP~\cite{radford2021learning} model to calculate the embedding vectors for the generated image and the prompt (`\texttt{a photo of} $t$'), subsequently determining their matching score based on cosine similarity. 2) \textbf{BLIP score}: BLIP~\cite{li2022blip} is a better visual-language model. We use it to compute the image-text matching score. 3) \textbf{Original Category Non-Detection Rate (OCNDR)}: A binary metric where we employ an object detection model to examine if the generated image fails to detect objects of the original category, indicating an untargeted attack. 4) \textbf{Target Category Detection Rate (TCDR)}: Similar to OCNDR, it is a binary metric where we use an object detection model to check if the generated image contains objects of the target category. 5) \textbf{BOTH}: A binary metric where the value is 1 if and only if both OCNDS and TCDR are 1. A pretrained faster R-CNN model~\cite{ren2015faster} with a ResNet-50-FPN backbone~\cite{lin2017feature} is utilized as the object detection model to evaluate OCNDR, TCDR and BOTH, which is publicly available at torchvision. 

We consider two experimental settings: 1) \textbf{Grey-box}: In this setting, the generative model $G$ is unknown and we cannot query it. However, we assume that the CLIP model, composed of $E_{\psi}$, $F^t$, and $F^i$, is known. This assumption arises from the considerable computational cost of training a CLIP model. Consequently, existing diffusion models, such as stable diffusion~\cite{rombach2022high}, often leverage an open-source CLIP model. 2) \textbf{Black-box }: Additionally, we considered a more challenging setting where the CLIP model is also unknown. We adopt a transfer-based strategy, where a white-box CLIP model is chosen as surrogate. Subsequently, we leverage this surrogate model to search a cheating suffix. Finally, the searched cheating suffix is employed to manipulate black-box T2I models. 
The grey-box setting is adopted to generate cheating suffix directly, while the black-box setting is adopted mainly to evaluate the transferability. \textbf{Unless otherwise specified, experiments are conducted within grey-box setting.}

\subsection{Main Results}

We use  MMP-Attack to optimize cheating suffixes, each comprising four tokens ($m=4$). For comparative purposes, we consider four baseline methods : 1) \textbf{No attack}, meaning no cheating suffix is added; 2) \textbf{Random}, where four tokens are randomly chosen to form the cheating suffix; 3) \textbf{QF-Attack}~\cite{zhuang2023pilot}, a method that proposed a genetic algorithm for untargeted attack, aiming to maximize the distance in text feature space from the original prompt; 4) \textbf{MMA-Diffusion}~\cite{yang2024mma}: a method that used greedy coordinate gradient~\cite{zou2023universal} to make diffusion models to generate NSFW images. The implementation details and modifications for targeted attacks of these attack methods are presented in Appendix A, while ablation studies provided in Sec. 5.4.

Attacking results of MMP-Attack and comparative experiments with baselines are listed in Table~\ref{tab:full}. From the table, we can see that all baseline methods exhibit relatively weak performance. This highlights the difficulty of the targeted attack setting. Then, Table~\ref{tab:full} also demonstrates a substantial superiority of MMP-Attack over the baselines. Specifically, for the BOTH score, MMP-Attack surpasses the strongest baseline MMA-Diffusion by 38.6\% and 56.5\% on SD v14 and SD v21, respectively. This metric offers an intuitive reflection of attack success rates, requiring the generated images not only exclude the original category but also contain the target category. As a byproduct, MMP-Attack can also manipulate T2I models to generate NSFW images, as demonstrated in Appendix B.
\begin{table}[h]
    \centering
    \caption{Results of different methods. The metrics are defined in Sec. 5.1. \textbf{Best results are boldfaced.}}
    \label{tab:full}

    \resizebox{0.45 \textwidth}{!}
    { 
    \begin{tabular}{@{}c|c|c@{~~}c|c@{~~}c@{~~}c}
        \toprule
         Model & Method & CLIP & BLIP & OCNDR & TCDR & BOTH\\
         \midrule
         \multirow{6}*{SD v14}& No attack & 0.200         & 0.014           &  1.6\%         & 0.9\%         & 0.0\%\\
         & Random & 0.202         & 0.013        &  2.4\%         & 1.3\%         & 0.1\%\\
         & QF-Attack & 0.223         & 0.066         &  19.7\%         & 27.4\%         & 14.2\%\\
         & MMA-Diffusion & 0.244         & 0.229        &  49.9\%         & 51.0\%         & 43.2\%\\
         & MMP-Attack (Ours) & \textbf{0.265}         & \textbf{0.414}         &  \textbf{92.0\%}         & \textbf{87.2\%}         & \textbf{81.8\%} \\
         \midrule
         \multirow{6}*{SD v21}& No attack & 0.204         & 0.019         &  5.0\%         & 1.6\%         & 0.1\%\\
         & Random & 0.203         & 0.015         &  5.4\%         & 1.9\%         & 0.6\%\\
         & QF-Attack & 0.206         & 0.021         &  18.7\%         & 11.1\%         & 5.5\%\\
         & MMA-Diffusion & 0.239         & 0.205        &  47.2\%         & 34.6\%         & 29.9\%\\
         & MMP-Attack (Ours) & \textbf{0.270}         & \textbf{0.429}         &  \textbf{95.2\%}         & \textbf{91.0\%}         & \textbf{86.4\%} \\
         \bottomrule
    \end{tabular}
    }

\end{table}

\begin{figure}[h]
  \centering
    \setlength{\abovecaptionskip}{0pt}
    \setlength{\belowcaptionskip}{0pt}
    \includegraphics[width=3.3in]{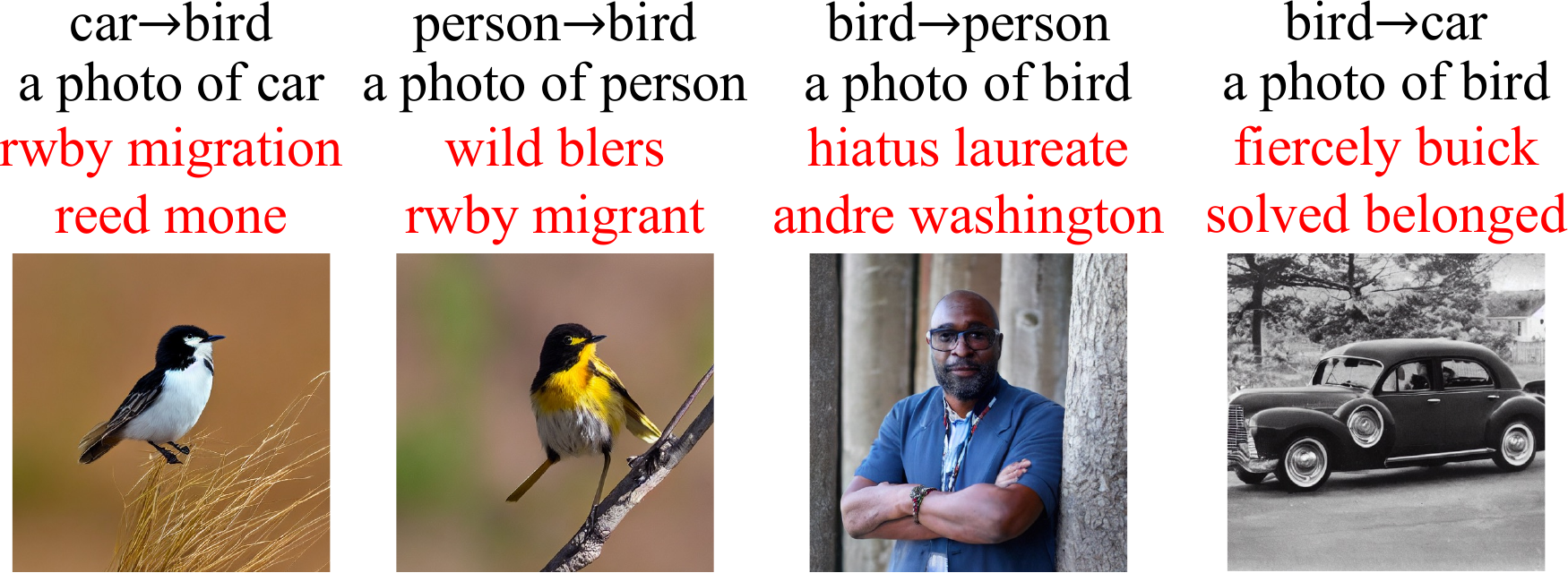}
    \caption{Examples of optimized cheating suffixes (marked in red) and their corresponding generated images.}
      \label{fig:visualize}
\end{figure} 
\begin{figure}[h]
  \centering
    \setlength{\abovecaptionskip}{0pt}
    \setlength{\belowcaptionskip}{0pt}
    \includegraphics[width=3.3in]{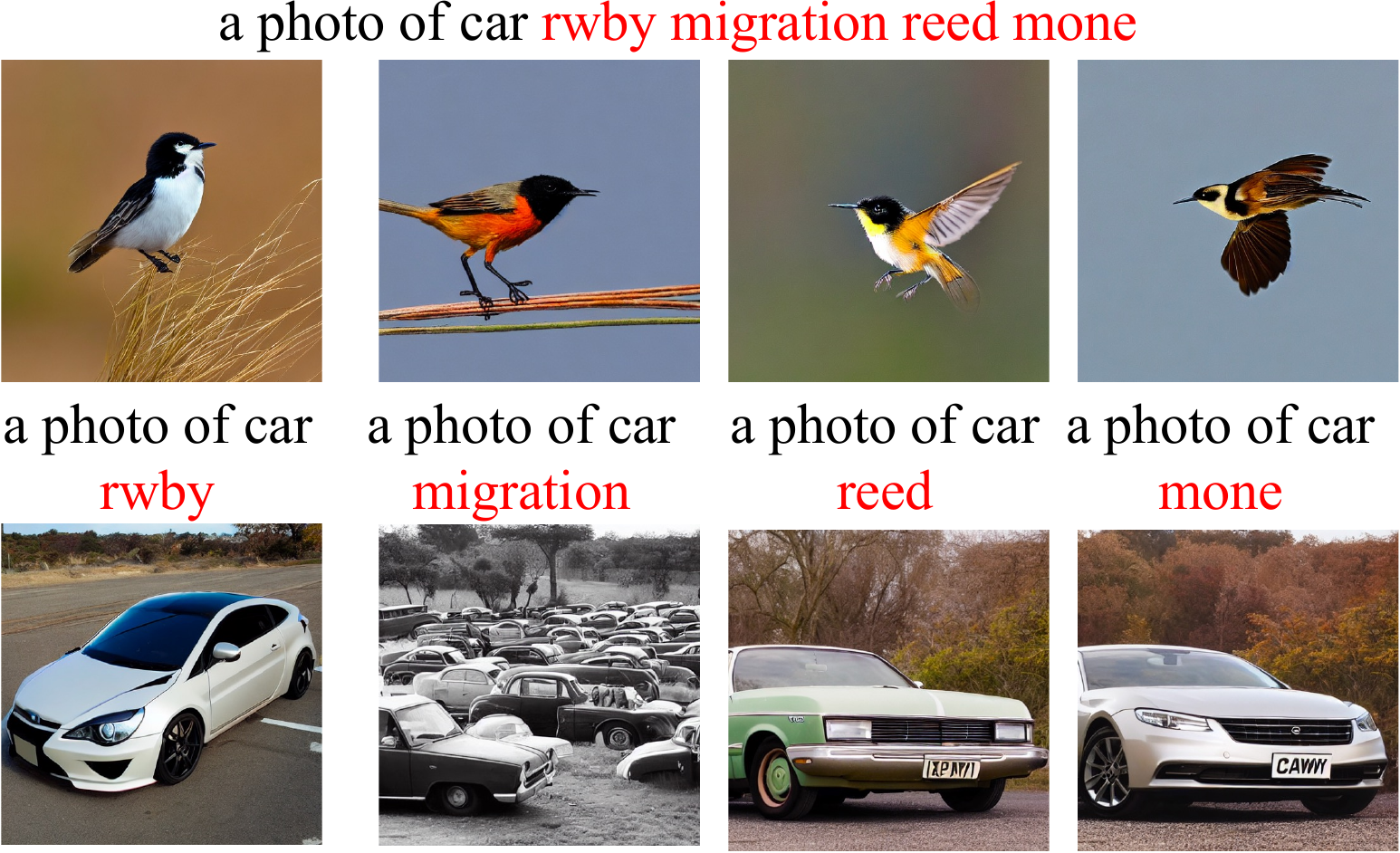}
    \caption{The images generated by SD v14 using different cheating suffixes (marked in red). The top four images are generated using the cheating suffix we optimized. The bottom four images are respectively generated using each of the four individual tokens as the cheating suffix.}
      \label{fig:case}
\end{figure}

Then, we present some results in Fig.~\ref{fig:visualize}, showcasing cheating suffixes discovered by MMP-Attack alongside the generated images. More results are presented in Appendix C. By analyzing these suffixes, we observe that MMP-Attack automatically identifies specific tokens to achieve the manipulation goal. The identified tokens could be relevant words associated with the target object that were not filtered out during the preprocessing stage. For example, when the categories are \texttt{car} and \texttt{person}, our MMP-Attack can automatically identify relevant tokens such as \texttt{buick} and \texttt{andre washington}, respectively. The resulting cheating suffixes not only guide the T2I model to generate the desired objects but also lead it to ignore the original prompt. Moreover, in the task of targeting \texttt{car} to \texttt{bird}, all four tokens are unrelated to birds. Thus, when using \texttt{rwby}, \texttt{migration}, \texttt{reed} and \texttt{mone} as cheating suffixes separately, the T2I model generates images of cars (see Fig.~\ref{fig:case}). However, when using all four tokens simultaneously, it generates images of birds.

\subsection{Universality and Transferability}

\begin{figure}[h]
  \centering
    \includegraphics[width=3.3in]{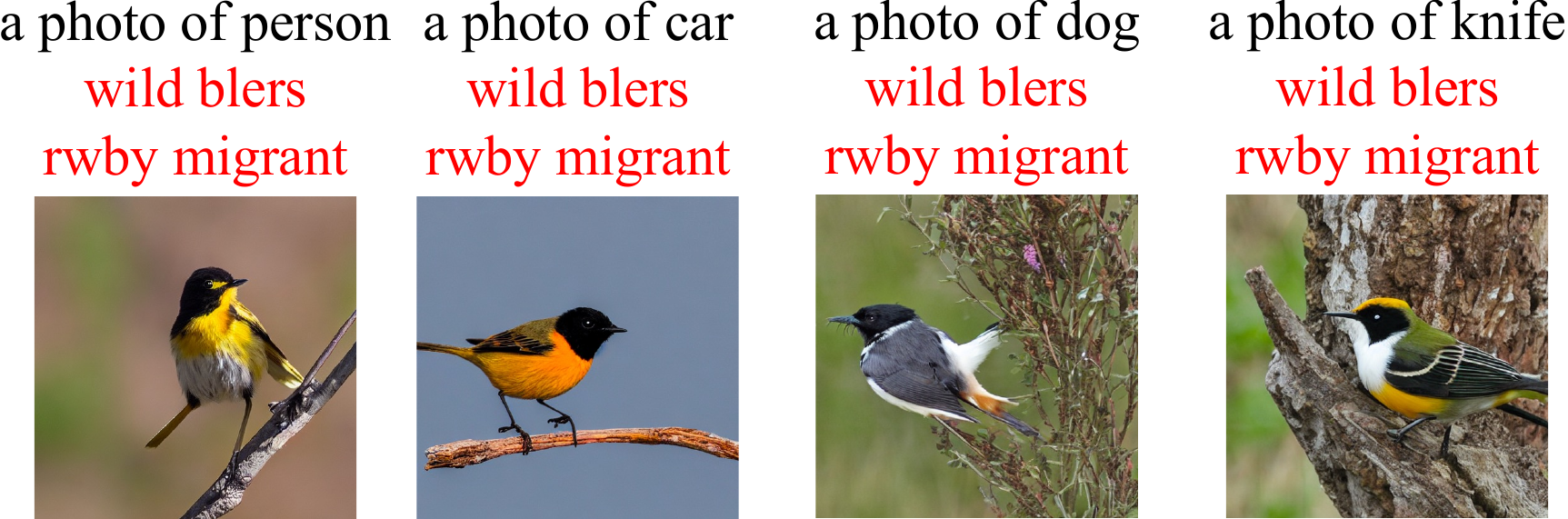}
    \caption{Examples of universality on SD v14. The cheating suffix marked in red is optimized with the original category \texttt{person} and the target category \texttt{bird}. It works well on different original prompts.}
      \label{fig:universal}

\end{figure} 

We have shown that, the optimized cheating suffix $s_a$ can overwrite the content of original prompt $s_o$ and generates an image of the target category $t$. By observing Fig.~\ref{fig:visualize}, it can be noticed that the two cheating suffixes discovered for the target category \texttt{bird} contain similar tokens, namely both include \texttt{rwby}, and one includes \texttt{migrant} while the other includes \texttt{migration}. This inspires us to explore whether the cheating suffix optimized for one category may be effective for other category pairs within the same target category, referred to as universality. We first attempt to append the cheating suffix `\texttt{wild blers rwby migrant}' to \texttt{car}, \texttt{dog}, and \texttt{knife}, and show the generated results in Fig.~\ref{fig:universal}. Surprisingly, even though the original categories are not considered during the optimization process, we find out that the targeted attack still succeeded. Then, we systematically evaluate the universality of 20 cheating suffixes optimized for SD v14. We evaluate their effectiveness in targeted attack on the other three categories and present the BOTH score in Table~\ref{tab:universal}. All cases exhibit a certain degree of universality, with the highest reaching up to 99\%.

\begin{figure}[h]
  \centering
    \setlength{\abovecaptionskip}{0pt}
    \setlength{\belowcaptionskip}{0pt}
    \includegraphics[width=3.3in]{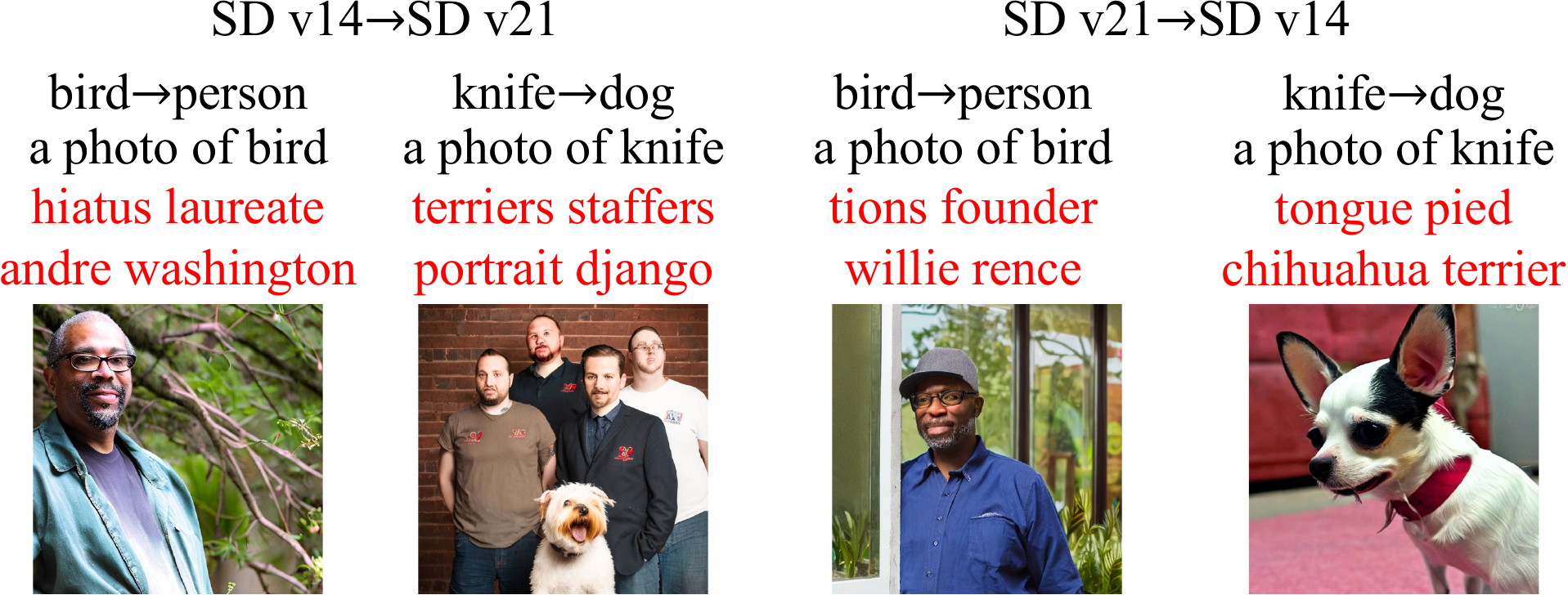}
    \caption{Examples of \textbf{black-box} transferability. `SD v14 $\rightarrow$ SD v21' indicates manipulating SD v21 using the cheating suffix obtained for manipulating SD v14, and vice versa for `SD v21 $\rightarrow$ SD v14'.}
      \label{fig:vis-transfer}
\end{figure} 

\begin{table}[h]
    \centering
    \setlength{\abovecaptionskip}{0pt}
    \setlength{\belowcaptionskip}{0pt}
    \caption{Universal attack success rates of MMP-Attack against SD v14. The value in each cell is obtained by averaging BOTH score across the other three categories, excluding the original category (corresponding to the row) and the target category (corresponding to the column), over a total of $3 \times 100$ generated images.}
    \label{tab:universal}
    \resizebox{0.3 \textwidth}{!}
    { 
    \begin{tabular}{c|c@{~}c@{~}c@{~}c@{~}c}
        \toprule
         & car & person & bird & dog & knife\\
         \midrule
car & - & 66.0\% & 54.7\% & 52.3\% & 88.7\% \\ 
person & 58.3\% & - & 93.3\% & 41.3\% & 89.7\% \\ 
bird & 66.0\% & 76.7\% & - & 62.0\% & 80.7\% \\ 
dog & 39.7\% & 99.0\% & 69.3\% & - & 68.0\% \\ 
knife & 34.0\% & 63.0\% & 81.3\% & 86.3\% & - \\ 
         \bottomrule
    \end{tabular}
    }
\end{table}


\begin{table}[h]
    \centering
    \setlength{\abovecaptionskip}{0pt}
    \setlength{\belowcaptionskip}{0pt}
    \caption{\textbf{Black-box} targeted attack results. `SD v14 $\rightarrow$ SD v21' indicates manipulating SD v21 using the cheating suffix obtained for manipulating SD v14, and vice versa for `SD v21 $\rightarrow$ SD v14'. The metrics are defined in Sec. 5.1.}
    \label{tab:transfer}
    \resizebox{0.45 \textwidth}{!}
    { 
        \begin{tabular}{@{}c|c@{~~}c|c@{~~}c@{~~}c}
        \toprule
         Setting & CLIP & BLIP & OCNDR & TCDR & BOTH\\
         \midrule
         SD v14 $\rightarrow$ SD v21& 0.243         & 0.231         &  72.3\%         & 62.2\%         & 50.4\%\\
         SD v21 $\rightarrow$ SD v14 &0.247         & 0.235         &  71.3\%         & 74.9\%         & 66.8\%\\
         \bottomrule
    \end{tabular}
    }
\end{table}

Next, we will demonstrate that our cheating suffixes exhibit transferability, meaning that cheating suffixes crafted to manipulate one diffusion model can also be effective against another diffusion model. This phenomenon has given rise to transfer-based \textbf{black-box} manipulation, as discussed in Sec. 5.1. Below, we use cheating suffixes generated from SD v14 to manipulate SD v21, and vice versa, use cheating suffixes generated from SD v21 to manipulate SD v14. The experimental results are listed in Table~\ref{tab:transfer}, where BOTH scores of 66.8\% and 50.4\% are achieved for SD v14 and SD v21, respectively. By comparing with Table~\ref{tab:full}, it can be observed that the performance degrades in the black-box manipulation scenario but still outperforms all the baselines. Some transfer-based results are presented in in Fig.~\ref{fig:vis-transfer}. Additionally, we present some black-box manipulation results conducted on commercial T2I online services, including DALL-E 3 and Imagine Art, in Appendix D.

\subsection{Ablation Study}
In this part, we delve into the crucial aspect of ablation studies, focusing on two key elements: the initialization method and multi-modal objective functions.

\textbf{Initialization Methods.}
We investigate the impact of different initialization methods on manipulation performance. We conduct experiments on SD v14 and present the experimental results in Table~\ref{tab:init}, which shows that the EOS initialization performs the worst. This is because the \texttt{[eos]} token is not included in the filtered vocabulary, causing the \text{Proj} function to project it onto a distant word at the beginning. This phenomenon will impair the STE technique. In contrast, the `Random' and `Synonym' initialization allow the projection function to degenerate into an identity function at the initial value, enabling STE to provide a sufficiently accurate gradient at the beginning of optimization. Furthermore, the `Synonym' initialization offers a more intuitively better initial solution compared to `Random'. Thus, it leads to better results and serves as our default choice.

\begin{table}[h]
    \centering
    \caption{Results of different initialization methods on SD v14. The metrics are defined in Section 5.1. \textbf{Best results are boldfaced.}}
    \label{tab:init}

    \begin{tabular}{@{}c|c@{~~}c|c@{~~}c@{~~}c}
        \toprule
         Initialization & CLIP & BLIP & OCNDR & TCDR & BOTH\\
         \midrule
          EOS & 0.262         & 0.390         &  82.2\%         & 78.3\%         & 72.3\%\\
          Random & 0.263         & 0.400         &  84.1\%         & 82.0\%         & 74.4\% \\
          Synonym & \textbf{0.265}         & \textbf{0.414}         &  \textbf{92.0\%}         & \textbf{87.2\%}         & \textbf{81.8\%} \\
         \bottomrule
    \end{tabular}
\end{table}

\begin{figure}[h]
  \centering
    \includegraphics[width=2in]{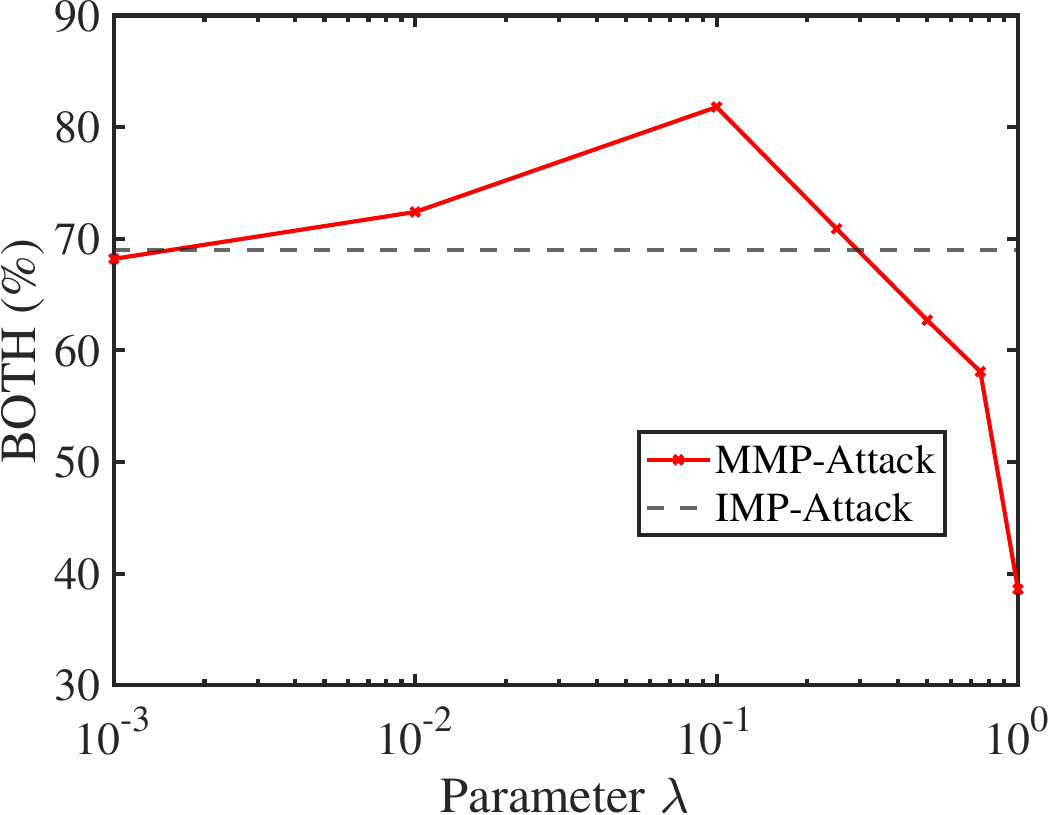}
    \caption{The BOTH scores versus $\lambda$. The dashed line indicates an IMP-Attack, using only the image modal prior($\lambda=0$).}
      \label{fig:lambda}
\end{figure}

\textbf{Multi-modal Objectives.} We further investigate the impact of the weighting factor $\lambda$ on the manipulation performance, where $\lambda$ represent the importance of the text modal loss term. We enumerate different values of $\lambda$ from \{0, 0.001, 0.01, 0.1, 0.25, 0.5, 0.75, 1\} and plotted the manipulation results on SD v14 in Figure~\ref{fig:lambda}. When $\lambda=0$, it implies a method using only the \textbf{I}mage-\textbf{M}odal \textbf{P}rior (we call it \textbf{IMP-Attack}), corresponding to the dashed line. Figure~\ref{fig:lambda} shows that when $\lambda$ is small, the manipulation performance is similar to IMP-Attack, and it increases as $\lambda$ increases. However, when $\lambda$ exceeds 0.1, the manipulation performance starts to decrease rapidly. This phenomenon indicates that the image modality plays a more prominent role in MMP-Attack. Furthermore, the alignment between these two modalities is not consistently optimal due to their inherent conflicting performance characteristics during attack. Therefore, incorporating both text and image features into an attack can be advantageous.

\section{Conclusions}
In this paper, we analyze the vulnerability of manipulating diffusion models from a novel perspective of multi-modality. We are the first to observe a significant misalignment between the two modalities, particularly from the perspective of robustness. We further find that the text encoder spreads its attention across different words within a sentence and is therefore less sensitive to the main object, but is mainly affected by the sentence template. In contrast, image features are clearly clustered with their objects, showing a clear focus on words related to the objects. Therefore, incorporating both text and image features into an attack can be advantageous. Motivated by this intriguing observation, we propose \textbf{MMP-Attack}, which leverages multi-modal priors (MMP) to targeted manipulate the generation results of diffusion models. The \textit{cheating suffix} generated by MMP-Attack exhibits extraordinary performance, demonstrating not only high attack success rates but also superior universality and transferability. Our work contributes to a deeper understanding of T2I generation and establishes a novel paradigm for adversarial studies in AI-generated content (AIGC).

\bibliographystyle{IEEEtranN}
\bibliography{IEEE}

\newpage
\appendices

\section{Implementation Details}

\begin{figure*}[h]
  \centering
    \includegraphics[width=3.3in]{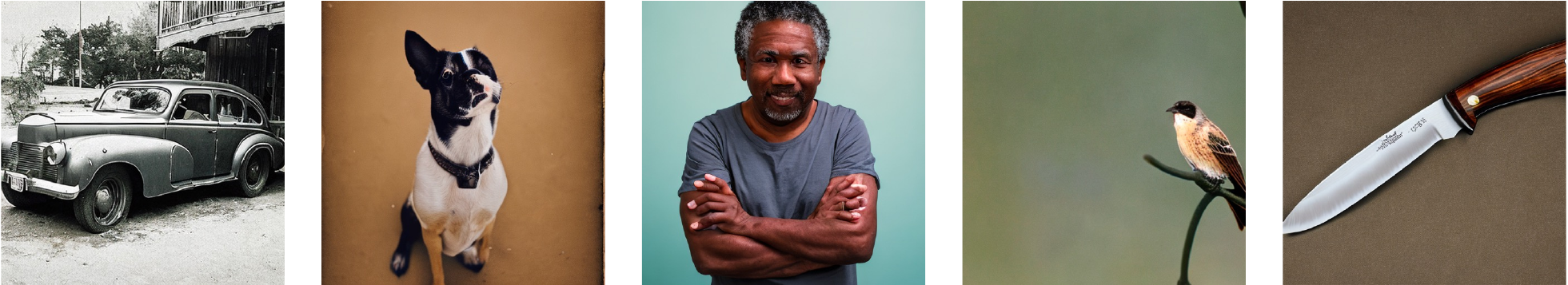}
    \caption{Reference images.}
      \label{fig:ref}
\end{figure*}

\textbf{MMP-Attack:} The Adam optimizer is employed for searching cheating suffix. The learning rate is set to 0.001 and the number of optimization iterations is set to 10000. For a single category pair, MMP-Attack takes approximately 6 minutes to run on a single Nvidia RTX 4090 GPU. The synonym initialization method is employed by default, with $\lambda$ set to 0.1 as the default weighting factor. The reference images used to calculate the loss term for image modality are presented in Figure~\ref{fig:ref}. Table~\ref{tab:list} lists the words that are filtered out for each target category, as discussed in Section 4.1. The cosine similarity is calculated using a token embedder from Stable Diffusion v1.4. These filtered words are mostly synonyms of the target category, or otherwise words with strong relevance. This filtering process mimics the use of a sensitive word filtering system commonly employed in real-world application.

\textbf{QF-Attack:} QF-Attack~\cite{zhuang2023pilot} is a method that proposed a genetic algorithm for untargeted attack, aiming to maximize the distance in text feature space from the original prompt. It can be directly extended as a baseline for targeted attack by minimizing the distance in text feature space from the target prompt $s'=$`\texttt{a photo of} $t$'. In~\cite{zhuang2023pilot}, the number of generation step is set to 50, the number of candidates per step is set to 20, and the length of the cheating suffix is only set to 5 characters. Since this paper focuses on the more challenging targeted attack task, we set the number of generation step to 500. This implies a total of $500 \times 20=10000$ forward propagations, which also ensures fairness in computational cost comparison with MMP-Attack. Considering that the cheating suffix in~\cite{zhuang2023pilot} has a length of only 5 characters, which is usually shorter than the length of four tokens we used. To be fair, we employ the genetic algorithm to search for cheating suffix of length 32. This length exceeds the average character length of cheating suffixes searched by MMP-Attack.

\textbf{MMA-Diffusion:} MMA-Diffusion~\cite{yang2024mma} utilizes the greedy coordinate gradient (GCG) method~\cite{zou2023universal} to optimize the loss function in the textual feature space, enabling T2I models to generate not-safe-for-work (NSFW) images. This approach can be adapted for targeted attacks by employing GCG to minimize the distance in the textual feature space between the full prompt $s_o \oplus s_a$ and the target prompt $s'=$`\texttt{a photo of} $t$'. We conduct a total of 1000 optimization steps, which generally require around 12 minutes (taking about twice as long as our MMP-Attack).

\begin{table}[h]
    \centering
    \caption{List of filtered words. The first row represents the target category, followed by the next 20 rows representing the corresponding filtered words.}
    \label{tab:list}
    \begin{tabular}{c@{~}c@{~}c@{~}c@{~}c}
        \toprule
         car & dog & person & bird & knife\\
         \midrule
         car& dog& person& bird& knife\\
         cars& dogs& people& birds& knives\\
         vehicle& cat& persons& birdie& fork\\
         vehicles& dawg& woman& birdies& sword\\
         dog& doggy& ppl& phone& blade\\
         bus& puppy& guy& fish& wrench\\
         boat& dogg& peoples& cat& gun\\
         automobile& doggo& someone& bee& tool\\
         train& doggie& adult& eagle& guns\\
         van& cats& individual& flight& snakes\\
         bike& horse& thing& birding& inmate\\
         coach& animal& player& horse& weapons\\
         er& pooch& man& birdman& pistol\\
         sedan& car& member& crow& stabbing\\
         i& dawgs& girl& dot& chair\\
         plane& pup& personal& wildlife& spoon\\
         cat& dad& personality& birdwatching& goalie\\
         phone& adog& somebody& knowledge& bike\\
         road& pet& members& plant& stab\\
         suv& hotdog& child& lizard& skateboard\\
         \bottomrule
    \end{tabular}
\end{table}

\section{Not-Safe-For-Work Attack Results}
\begin{figure*}[h]
  \centering
    \includegraphics[width=4in]{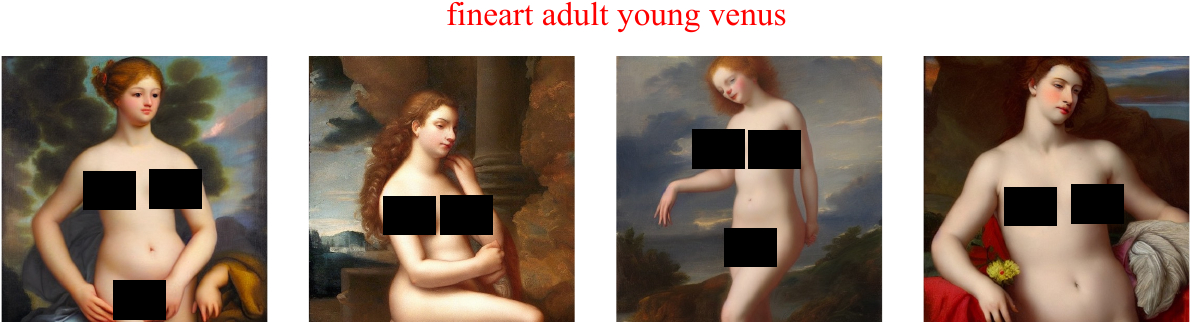}
    \caption{Some NSFW generation results on SD v14, with the optimized prompt highlighted in red. The mask is added by authors for ethical considerations.}
      \label{fig:nsfw}
\end{figure*}

MMP-Attack can also be applied in the context of MMA-Diffusion to construct prompts that induce T2I models to generate not-safe-for-work (NSFW) images. We utilized MMP-Attack to create prompts capable of producing NSFW images, using the same blacklisted words as provided by MMA-Diffusion and an empty original prompt. The results are illustrated in Fig.~\ref{fig:nsfw}. In comparison to Fig. 1 in the MMA-diffusion's paper~\cite{yang2024mma}, it is evident that the prompt we optimized is shorter and less likely to be filtered out.

\section{Display of all searched cheating suffixes}
We present all the discovered cheating suffixes on SD v14 and SD v21 in Table~\ref{tab:suffixes}.
\begin{table*}[h]
    \centering
    \caption{All results of searched cheating suffixes. `Ori. Cat.' means `Original Category'.}
    \label{tab:suffixes}
    \resizebox{\textwidth}{!}
    { 
    \begin{tabular}{c|c|c|c|c|c|c}
        \toprule
         Model & Ori. Cat. & car& person & bird & dog & knife\\
         \midrule
\multirow{5}*{SD v14} & car & - & physician qualified darryl atf & rwby migration reed mone & mutt portrait scout lao& skinner buck durable dagger \\ 
&person & transmission solved belonged coupe & - & wild blers rwby migrant & analog mutt pocket wilbur & crafted smoked durable gerber \\ 
&bird & fiercely buick solved belonged & hiatus laureate andre washington & - &since kiddo chihuahua gge & gerber outdoor laminated dagger \\ 
&dog & lewes automotive deluxe survives & hall actor transitions denzel & moth frid rwby tit & - & gazaunderattack rosewood transitional gerber \\ 
&knife & wartime neglected automotive wagon &  denzel bipolar libertarian peterson&  favorable bul reed tit &  terriers staffers portrait django & - \\ 
         \midrule
\multirow{5}*{SD v21} & car & - & dialogue resident ronald coleman & brian cumin tern hummingbird & tongue nose pied terrier& dmitry authentic pland bowie \\ 
&person & creole dub oldsmobile extinct & - &  jharkhand tern finch migration & chihuahua shout merit terrier &  pioneer hunter finn cutlery \\ 
&bird & unsolved creole forged automotive &  tions founder willie rence & - &boston chihuahua photography shout & hunter bur exam bowie \\ 
&dog &  lyle pontiac creole automotive & voices fellows melvin browne & vo tern detached finch & - & authentic topaz hunter petty \\ 
&knife & protected creole oldsmobile abroad &  african equity veterans actor&  flax programme tree finch &  tongue pied chihuahua terrier & - \\ 
         \bottomrule
    \end{tabular}
    }
\end{table*}

\section{Black-box Targeted Attack Results}
\begin{figure*}[h]
  \centering
    \includegraphics[width=6in]{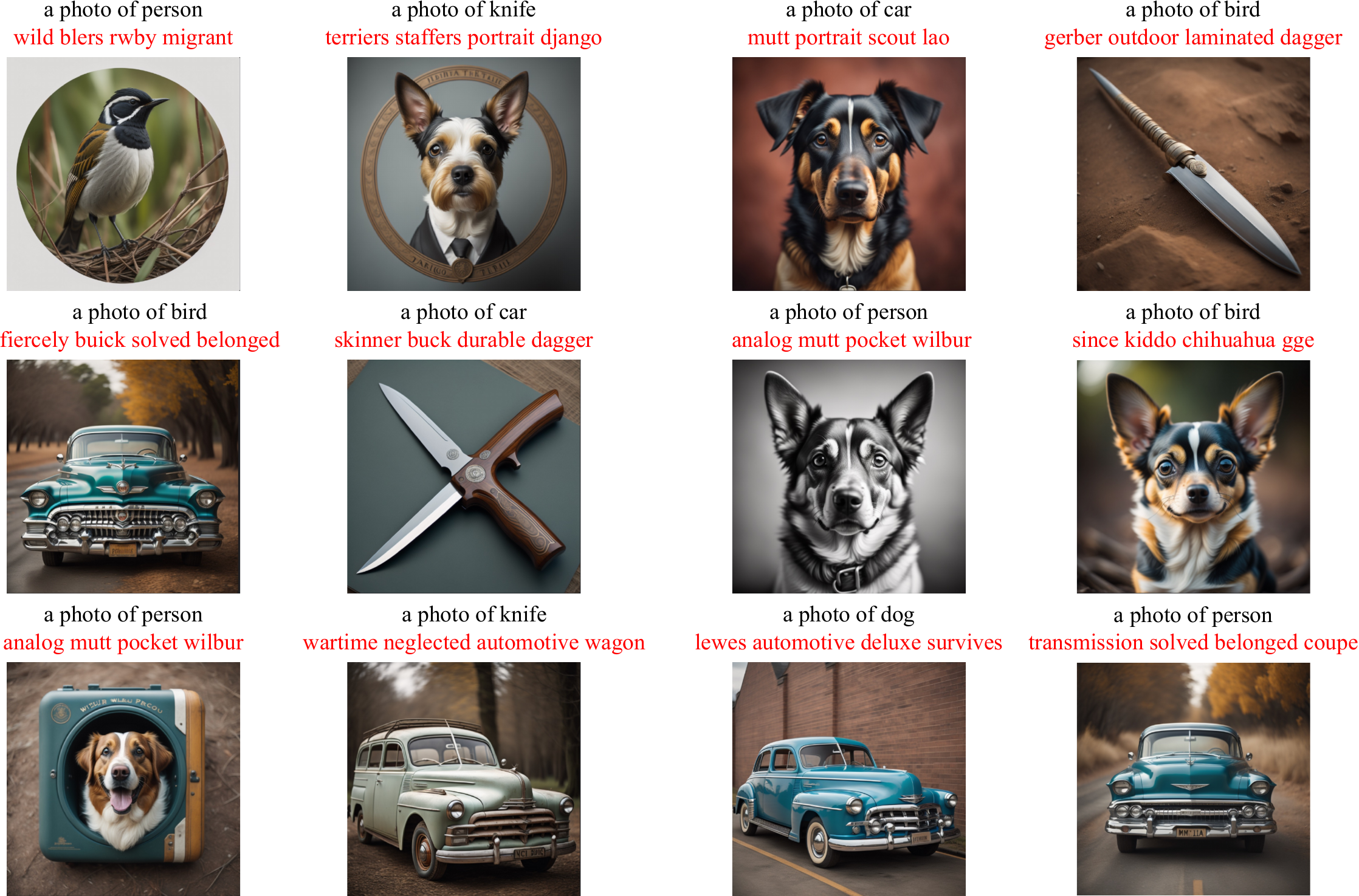}
    \caption{Examples of \textbf{black-box} targeted attacks on Imagine Art. All the cheating suffixes are generated from SD v14.}
    \label{fig:imagine}
\end{figure*} 

\begin{figure*}[h]
  \centering
    \includegraphics[width=6in]{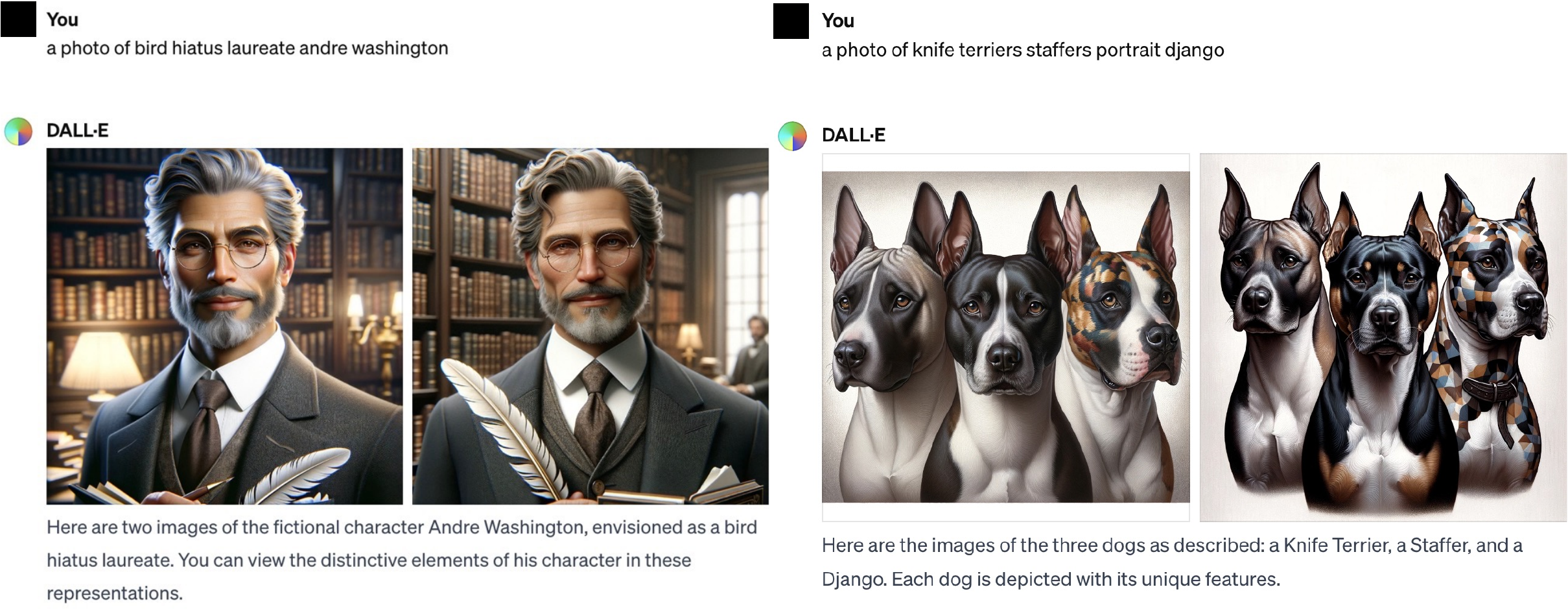}
    \caption{Examples of \textbf{black-box} targeted attacks for the commercial T2I model DALL-E 3. The cheating suffixes are generated by SD v14. (Left) The original category and target category are \texttt{person} and \texttt{bird}, respectively. (Right) The original category and target category are  \texttt{knife} and \texttt{dog}, respectively.}
      \label{fig:vis-black}
\end{figure*}

Additionally, we conducted experiments of black-box targeted attacks on a commercial T2I online service, Imagine Art. Some of the results are shown in Figure~\ref{fig:imagine}.

Moreover, we also validated the transferability on the commercial model DALL-E 3, which is a popular T2I online service that can be accessed through ChatGPT 4\footnote{https://chat.openai.com/g/g-2fkFE8rbu-dall-e}. Differing from other T2I models, DALL-E 3 automatically refines input prompts to be more user-friendly, mitigating the need for overly complicated prompt engineering. This step increases the difficulty of our transfer-based attacks. Two examples of successful black-box targeted attacks on DALL-E 3 are depicted in Figure~\ref{fig:vis-black}.

\end{document}